%% file: main.tex
\documentclass{llncs}

\input{packages}

\input{macros}

\begin{document}
\title{Spectral Evolution with Approximated Eigenvalue Trajectories for
  Link Prediction}
\titlerunning{Spectral Analysis for Link Prediction in Complex Networks}
\author{Miguel Romero \and Jorge Finke \\ Camilo Rocha  \and Luis Tob{\'o}n}
\authorrunning{Miguel Romero et al.}

\institute{Department of Electronics and Computer Science\\
  Pontificia Universidad Javeriana, Cali, Colombia.}

\maketitle              % typeset the title of the contribution

\input{abstract}

\input{intro}

\input{model}

\input{extmodel}

\input{results}

\input{concl}

\bibliographystyle{apa-good}
\bibliography{miguel,jorge,camilo}
\end{document}

%% file: packages.tex
\usepackage{url}
\usepackage{natbib}
\usepackage{amsmath}
\usepackage{amssymb}
\usepackage{amsmath}
\usepackage{numprint}
\usepackage{tabularx}
\usepackage{multirow}
\usepackage{graphicx}
\usepackage{todonotes}
\usepackage{subcaption}
\usepackage[export]{adjustbox}

%% file: macros.tex
\newcolumntype{L}{>{\raggedright\arraybackslash}X}
\newcolumntype{R}{>{\raggedleft\arraybackslash}X}
\newcolumntype{Y}{>{\centering\arraybackslash}X}

\newcommand{\MAT}[1]{\mathbf{#1}}
\newcommand{\nats}{\mathbb{N}}

%% file: abstract.tex
\begin{abstract}
  The spectral evolution model aims to characterize the growth of
  large networks (i.e., how they evolve as new edges are established)
  in terms of the eigenvalue decomposition of the adjacency
  matrices.  It assumes that, while eigenvectors remain constant,
  eigenvalues evolve in a predictable manner over time. This paper
  extends the original formulation of the model twofold.
  First, it presents a method to compute an approximation of the
  spectral evolution of eigenvalues based on the Rayleigh quotient.
  Second, it proposes an algorithm to estimate the
  evolution of eigenvalues by extrapolating only a fraction
  of their approximated values.
  The proposed model is used to characterize mention networks of users
  who posted tweets that include the most popular political hashtags
  in Colombia from August $2017$ to August $2018$ (the period which
  concludes the disarmament of the Revolutionary Armed Forces of
  Colombia). To evaluate the extent to which the spectral evolution
  model resembles these networks, link prediction methods based on
  learning algorithms (i.e., extrapolation and regression) and graph
  kernels are implemented. Experimental results show that the learning
  algorithms deployed on the approximated trajectories outperform the
  usual kernel and extrapolation methods at predicting the formation
  of new edges.

\keywords{link prediction, spectral evolution model, Twitter mention
  networks, spectral decomposition, Rayleigh quotient, graph kernels}
\end{abstract}

%% file: intro.tex
\section{Introduction}
\label{sec:intro}

% why
Many systems of interest have the form of a \textit{network} (or
\textit{graph}) where entities are represented as \textit{nodes} and
relationships between pairs of nodes as \textit{edges}. Modeling and
analyzing these networks tend to become more complex as they grow in
size (both in terms of nodes and edges) and in the number of
concurrent processes (activities that occur at the same time across
different nodes). Not surprisingly, efforts to explain the tangled
dynamics of empirical networks of overwhelming size have received
increasing attention in recent years. Numerous approaches have been
proposed to gain a better understanding of network behavior. For
example, the detection of malicious accounts in location-based social
networks is a first step to provide reliable information to users and
improve their user experiences~\citep{gong-deepscan-2018}. Predicting
the relationships of online social networks has been key in
identifying whether influential users maintain their status, or
whether a political polarization of a group of nodes reflects a
transitory or stationary state~\citep{dimaggio-polarized-1996}.
Similarly, information of users across different social platforms has
been used to develop a multi-layer algorithm that predicts links on
social networks~\citep{jalili-multilayer-2017}.  Modeling
Twitter data as a co-occurrence language network has also been used
to infer future relationships between
users~\citep{martincic-twitter-2017}.  In all these studies, a common
goal is to develop models that capture node interactions and predict
how new edges are established over time.

The field of ~\textit{spectral graph theory} is one of the fundamental
pillars for the study of networks~\citep{godsil-agt-2001}. It uses
eigenvalues and eigenvectors of matrices associated with graphs to
shed light and reveal combinatorial properties of
networks~\citep{kurucz-spectclust-09}. Common associations include the
adjacency and Laplacian matrices, which range over the nodes of the
graph and may be symmetric or asymmetric. In either case, such
matrices can be viewed both as an operator mapping vectors to vectors
and as a mean to define quadratic forms. Any real-valued symmetric
matrix $\mathbf{A}$ with $n$ rows (and columns) can be characterized
by its \emph{spectral decomposition}, i.e., a set of pairs of
eigenvalues and eigenvectors
$(\MAT{X}, \Lambda) = \{(\mathbf{x_1}, \lambda_1), \ldots,
(\mathbf{x_n}, \lambda_n)\}$
such that the equation $\mathbf{A}\mathbf{x_i} =
\lambda_i\mathbf{x_i}$ holds for each $1 \leq i \leq n$. Therefore,
the evolution of a graph with $n$ vertices can be represented either by a sequence $\mathbf{A}_1, \ldots, \mathbf{A}_t$ of symmetric $n
\times n$ matrices or by a sequence of their spectral decompositions
$(\MAT{X}_1, \Lambda_1), \ldots, (\MAT{X}_t, \Lambda_t)$.
From a practical viewpoint, spectral decomposition becomes a powerful
tool to predict, under certain assumptions on how each pair of
eigenvalues and eigenvectors behaves, the evolution of networks.

In particular, the \textit{spectral evolution
  model}~\citep{kunegis-semodel-2013} characterizes the dynamics of a
network (i.e., how new edges are created over time) in terms of the
evolution of its spectra, assuming that the associated eigenvectors
remain largely constant. That is, within the spectral evolution model
framework, each sequence of symmetric matrices $\mathbf{A}_1, \ldots,
\mathbf{A}_t$ is characterized solely by the set of eigenvectors, say
$\MAT{X}_t$, and the spectra $\Lambda_1, \ldots, \Lambda_t$. By
neglecting possible minor changes in the behavior of the eigenvectors,
efforts to characterize the dynamics of a network focus exclusively on
predicting the trajectory of the eigenvalues.

% what
This paper extends the original  spectral evolution
model~\citep{kunegis-semodel-2013} by introducing a tandem of two
techniques that complement each other.
%how
%
First, the extended model predicts how a network grows based on
extrapolating an approximation of its eigenvalues.  Common kernel and
extrapolation methods are memoryless in the sense that only one or two
network snapshots are considered for extrapolating eigenvalues.  In
other words, the evolution history of the network is mostly neglected.
The working hypothesis is that, by considering an \textit{entire}
sequence of eigenvalue approximations $\hat{\Lambda}_1, \ldots,
\hat{\Lambda}_t$, better eigenvalue trajectories may be computed to
predict links at time $t+1$. Each approximation $\hat{\Lambda}_i$ is
computed by using the Rayleigh
quotient~\citep{chatelin-eigenvalues-2012} and extrapolation can be
achieved by fitting the trajectory of the approximated eigenvalues
with regression algorithms. The resulting approach to compute
$\MAT{A}_{t+1}$ takes $O(tn^3)$ time, which is asymptotically
equivalent to checking the extended model hypothesis on the sequence
$\mathbf{A}_1, \ldots, \mathbf{A}_t$ with state-of-the-art
algorithms. In practice, this means that the proposed approach can
always be used as an extension of the spectral evolution model,
without adding significant computational costs.

The second contribution is based on an experimental observation: The learning 
algorithms can accurately predict the evolution of
the eigenvalues by only taking into account a fraction of all the approximated
trajectories. That is, instead of working with the $n$ eigenvalue
trajectories to compute $\MAT{A}_{t+1}$, it is sufficient to consider a fraction 
of the trajectories in order to achieve an accurate link
prediction. Therefore, the above-mentioned computational effort  can be
reduced further in practice by a constant factor. For instance, in
some experiments, the adjacency matrix $\MAT{A}_{t+1}$ can be
recovered with an accuracy of $99.97\%$ when only $8\%$ of the
trajectories are considered.
%
%These two contributions complement existing methods based on the
%spectral evolution model for link prediction in complex networks.

% case study
The proposed approach is used to explain the evolution networks of
Twitter users who posted messages with political hashtags. In these
networks, each node represents a user. An edge is present between
two users (nodes) whenever any of the two users mentions the other in
a message containing a selected hashtag. The selection corresponds to
the most popular hashtags associated to political affairs in Colombia
between August $2017$ and August $2018$.  This period marks the
disarmament of the Revolutionary Armed Forces of Colombia (Farc) and
the end of an armed conflict that lasted over 50
years~\citep{ince-farc-2013,saab-farc-2009}. The experimental
exploration suggests that spectral extrapolation methods based on
linear and quadratic regression outperform the approaches based on
graph kernels. Achieving a better performance is related to the
ability to capture the irregular evolution of the eigenvalues (i.e., 
for scenarios in which the rate of growth of some eigenvalues is higher than for others).
% and
%to the fact that entire spectral trajectories are being considered for
%extrapolation.

% difference with conference paper
Finally, it is important to note that this paper builds on results
presented in~\citep{romero-sem-2020}. Unlike the work
in~\citep{romero-sem-2020}, the model
presented here is equipped with the Raleigh quotient for computing
approximated spectra. Furthermore, new learning algorithms have been
evaluated to predict the trajectories of the eigenvalues. Finally,
unlike~\citep{romero-sem-2020}, the mention networks are not
considered multi-graphs but simple graphs, which renders edge
prediction into a binary classification problem.

% outline
\textbf{Outline.} The remainder of the paper is organized as
follows. Section~\ref{sec:model} presents the spectral evolution
model. Section~\ref{sec:extmodel} introduces the extended spectral
evolution model, including the Rayleigh quotient and the approximation
of spectral trajectories.  Section~\ref{sec:results} describes the
mention networks used for the experimental analysis and presents the
results of applying the proposed approach to these networks.
Section~\ref{sec:concl} draws some concluding remarks and future
research directions.

%% file: model.tex
\section{Preliminaries}
\label{sec:model}

This section presents preliminaries on the spectral evolution model,
including common growth methods based on graph kernels and spectral
extrapolation used for link prediction. The overview is, mainly, based
on~\citep{kunegis-semodel-2013}.

\subsection{Spectral Representation}

A graph $G = (V, E)$ is specified by its \textit{vertex} set $V$ and
its \textit{edge} set $E$; throughout the rest of paper it is assumed
that any graph is finite (i.e., $|V|$ is a natural number),
undirected, and without self or multiple edges. Without loss of
generality, the vertex set $V$ is assumed to be ordered, e.g., $V =
\{1, 2, 3, \ldots, |V|\}$.
To any graph $G = (V, E)$ a matrix $\MAT{A}_G$ (or simply $\MAT{A}$)
is associated whose entries are given for any $u,v \in V$ by
$\MAT{A}(u, v) = 1$ if $(u, v) \in E$ and $\MAT{A}(u, v) = 0$ if $(u,
v) \notin E$. The matrix $\MAT{A}$ is called the \textit{adjacency
  matrix} of $G$. Under the above assumptions, any adjacency matrix is
symmetric.

A vector $\MAT{x}$ is an \textit{eigenvector} of a matrix $\MAT{A}$
with \textit{eigenvalue} $\lambda$ iff $\MAT{A}\MAT{x} =
\lambda\MAT{x}$. The spectral theorem states an important fact about
symmetric adjacency matrices.

\begin{theorem}\label{thm:eigendecom}
  Let $n \in \nats$ be positive. If $\MAT{A}$ is a $n$-by-$n$
  symmetric matrix, with real entries, then there exist real numbers
  $\lambda_1, \lambda_2, \ldots, \lambda_n$ and $n$ mutually
  orthogonal unit vectors $\MAT{x}_1, \MAT{x}_2, \ldots, \MAT{x}_n$
  such that $\MAT{x}_i$ is an eigenvector of $\MAT{A}$ with eigenvalue
  $\lambda_i$, for $1 \leq i \leq n$.
\end{theorem}

The pair $(\MAT{X}, \MAT{\Lambda})$, with $\MAT{X}$ being the
orthogonal matrix obtained from the unit vectors
$\MAT{x}_1,\ldots,\MAT{x}_n$ and $\MAT{\Lambda}$ the diagonal matrix
determined by $\lambda_1,\ldots, \lambda_n$ in
Theorem~\ref{thm:eigendecom}, is called the \emph{spectral
  decomposition} of $\MAT{A}$ and satisfies $\MAT{A} =
\MAT{X}\MAT{\Lambda}\MAT{X}^T$. It is important to note that
eigenvalues are uniquely determined, but can be repeated. However,
eigenvectors are not uniquely determined for a given eigenvalue. In
general, the eigenvectors of a given eigenvalue are determined modulo
orthogonal transformations.

\subsection{The Spectral Evolution Model}

The \textit{spectral evolution model}~\citep{kunegis-semodel-2013}
assumes that the evolution of a network can be described by the
evolution in the network's spectrum, while the networks eigenvectors
stay largely constant over time.
Formally, if $\MAT{A}_t$ is the adjacency matrix of a network at time
$t$ and $(\MAT{X}_t, \MAT{\Lambda}_t)$ its spectral decomposition,
then the spectral decomposition $(\MAT{X}_{t'}, \MAT{\Lambda}_{t'})$
of the adjacency matrix of the network at any time $t'$ satisfies
$\MAT{X}_{t} \approx \MAT{X}_{t'}$ (i.e., the eigenvectors of
$\MAT{A}_t$ and $\MAT{A}_{t'}$ are approximately the same over time).

There are, basically, two existing approaches to link prediction with
the spectral evolution model: spectral transformation functions (based on graph kernels) and
spectral extrapolation.

\subsubsection{Spectral Transformation}

Link prediction can be expressed as a spectral decomposition
$\MAT{X}F(\MAT{\Lambda})\MAT{X}^T$, where $F$ is a \textit{spectral
  transformation} that applies the same real function to each diagonal
element $\lambda_j$ of $\MAT{\Lambda}$.
%
%Graph kernels are spectral transformation functions commonly used for
%link prediction in combination with the spectral evolution model.

\textbf{The Triangle Closing Kernel.}
The triangle closing kernel is expressed as
$\MAT{A}^2=\MAT{X}\MAT{\Lambda}^2\MAT{X}^T$. That is, the kernel
replaces the eigenvalues of
$\MAT{A}$ by their squared values. The  function associated to
the triangle closing kernel is $f(\lambda)=\lambda^2$.

\textbf{Exponential Kernel.}
The exponential kernel is expressed as
\[\exp{(\alpha\MAT{A})}=\sum_{k=0}^\infty\alpha^k\frac{1}{k!}\MAT{A}^k,\]
where $\alpha$ is a constant used to balance the weight of short and
long paths. It denotes the sum of every path between two vertices
weighted by the inverse factorial of its length. The function
associated to the exponential kernel is
$f(\lambda)=e^{\alpha\lambda}$.

\textbf{Neumann Kernel.}
The Neumann kernel is expressed as
\[(\MAT{I}-\alpha\MAT{A})^{-1}=\sum_{k=0}^\infty\alpha^k\MAT{A}^k,\]
where $\alpha^{-1}>|\lambda_1|$ and $\lambda_1$ is the largest
eigenvalue of $\MAT{A}$. The function associated to the Neumann kernel is
$f(\lambda)=1/(1-\alpha\lambda)$.

\subsubsection{Spectral Extrapolation}

Graph kernels can characterize a spectral transformation function for
the growth of the spectrum. However, when the evolution of the
spectrum is irregular, it is not always possible to find a simple
function that describes the trajectory for all eigenvalues. In this
situation, it is often convenient to use spectral extrapolation, which
can be viewed as a generalization of graph kernels.

Let $t_1$ be an intermediate time and $t_2$ the final time from which
link prediction is to be made, and let $\MAT{A}_1$ and $\MAT{A}_2$ be
the adjacency matrices of the network at times $t_1$ and $t_2$,
respectively. These matrices can be decomposed as $\MAT{A}_1 =
\MAT{X}_1\MAT{\Lambda}_1\MAT{X}_1^T$ and $\MAT{A}_2 =
\MAT{X}_2\MAT{\Lambda}_2\MAT{X}_2^T$, for some
$\MAT{X}_1,\MAT{\Lambda}_1,\MAT{X}_2,\MAT{\Lambda}_2$. If
$(\lambda_{2})_j$ is the $j$-th eigenvalue (say, under the ordering of
$\MAT{\Lambda}_2$'s diagonal) at $t_2$, its estimated previous value
$(\hat{\lambda}_1)_j$ at $t_1$ is computed as a diagonalization of
$\MAT{A}_1$ by $\MAT{X}_2$
\[(\hat{\lambda}_1)_j=\left(\sum_i(\MAT{X}_{1})_i^T(\MAT{X}_2)_j\right)^{-1}\sum_i(\MAT{X}_1^T)_i(\MAT{X}_2)_j(\lambda_1)_i,\]
where $(\MAT{X}_t)_k$ is the $k$-th eigenvector for $1 \leq k \leq n$,
and $(\lambda_t)_k$ is the $k$-th eigenvalue of $\MAT{A}_t$ ($t \in
\{1,2\}$), respectively (under the same ordering mentioned
above). Note that in this formula, the occurrences of $\MAT{X}_{1}$
could be replaced by $\MAT{X}_{2}$ because of the spectral evolution
model assumptions. However, if the possible small changes in the
eigenvectors are to be taken into account, the given version of the
formula should be used.

Linear extrapolation is used to predict link formation in the network
at a future time $t_3>t_2$. The $j$-th eigenvalue at $t_3$ is
approximated from the eigenvalues $\lambda_2$ at time $t_2$ and their
approximated counter-parts $\hat{\lambda}_1$ at time $t_1$, as
follows:
\[(\hat{\lambda}_3)_j=2(\lambda_{2})_j-(\hat{\lambda}_1)_j.\]
Finally, a link prediction real-valued matrix $\hat{\MAT{A}}_3$ is
computed as follows:
\[\hat{\MAT{A}}_3=\MAT{X}_2\hat{\MAT{\Lambda}}_3\MAT{X}_2^T.\]
Let $\hat{\MAT{A}}_3(u, v)$ denote the entry in row $u$ and column $v$
of $\hat{\MAT{A}}_3$.  For each $u,v \in V$, the value
$\hat{\MAT{A}}_3(u, v)$ is a link prediction score representing the
probability of having at time $t_3$ a link between $u$ and $v$. The
adjacency matrix $\MAT{A}_3$, which would represent the network at the
future time $t_3$, can be constructed from $\hat{\MAT{A}}_3$ in many
ways. For example, by setting a threshold $0 \leq \delta \leq 1$ such
that $\MAT{A}_3(u,v)=1$ if $\hat{\MAT{A}}_3(u,v) \geq \delta$ and
$\MAT{A}_3(u,v)=0$ if $\hat{\MAT{A}}_3(u,v) < \delta$, for each $u,v
\in V$.

%% file: extmodel.tex
\section{Extending the Spectral Evolution Model}
\label{sec:extmodel}

Consider a network $G$ with $n$ vertices and a sequence of
($n$-by-$n$) adjacency matrices $\MAT{A}_1, \MAT{A}_2, \ldots,
\MAT{A}_t$ representing the evolution of $G$ (i.e., the addition of
new edges among existing nodes) over $t$ time units. One or more edges
can be created from one time unit to the next. The problem of
predicting the formation in $G$ of new edges at time $t+1$ consists in
extending the given sequence of adjacency matrices with a new
adjacency matrix $\MAT{A}_{t+1}$.  The spectral evolution model
introduced in Section~\ref{sec:model} can be used for such a task in
two ways, under the assumption that the eigenvectors of the adjacency
matrices remain constant. On the one hand, a graph kernel can be
chosen as a specific spectral transformation function for obtaining
$\MAT{A}_{t+1}$ from $\MAT{A}_t$. On the other hand, a time $t' < t$
can be picked so that $\MAT{A}_{t+1}$ is extrapolated from
$\MAT{A}_{t'}$ and $\MAT{A}_t$.  In either case, the evolution history
of the network $G$ is not fully considered and mostly neglected. The
working hypothesis is that, by considering the \textit{entire}
sequence $\MAT{A}_1, \MAT{A}_2, \ldots, \MAT{A}_t$, better eigenvalue
predictions could be computed to build $\MAT{A}_{t+1}$. That is, the
more information about how eigenvalues change over the $t$ time units,
the better link prediction for the network would be at time $t+1$. Of
course, the key question to ask is how to use all this information and
yet have an efficient construction method for $\MAT{A}_{t+1}$: this
section introduces an extrapolation-based approach that answers this
question by using approximations computed with the Rayleigh quotient.

\subsection{The Rayleigh Quotient}
\label{sec:rayleigh}

The eigenvalues and eigenvectors of symmetric matrices have many
characterizations. The Rayleigh quotient is a tool from linear algebra
that offers one of such characterizations (see, e.g.,
~\citep{chatelin-eigenvalues-2012}).

\begin{definition}\label{def:rayleigh}
  Let $n \in \nats$ be positive, $\MAT{A}$ an (adjacency) $n$-by-$n$
  matrix, and $\MAT{x}$ a real-valued non-zero $n$-vector. The
  \textit{Rayleigh quotient} of $\MAT{x}$ with respect to $\MAT{A}$ is
  the ratio $R(\MAT{A}, \MAT{x})$ defined by
  \[R(\MAT{A}, \MAT{x}) = \frac{\MAT{x}^T\MAT{A}\MAT{x}}{\MAT{x}^T\MAT{x}}.\]
\end{definition}
Note that the Rayleigh quotient $R(\MAT{A}, \MAT{x})$ is a real number
(i.e., a scalar value). If $\MAT{x}$ is an eigenvector of $\MAT{A}$,
then the ratio $R(\MAT{A}, \MAT{x})$ is the corresponding
eigenvalue.
%% Also, if $\MAT{x}$ is a vector that maximizes $R(\MAT{A},
%% \MAT{x})$, then $\MAT{x}$ is an eigenvector of $\MAT{A}$ (and the
%% one which associated eigenvalue is largest among all eigenvalues of
%% $\MAT{A}$). Moreover, such a vector is guaranteed to exists. This
%% is because it suffices to consider unit vectors $\MAT{x}$ and this
%% set is compact: the Rayleigh quotient is a continuous function away
%% from the origin and, thus, there exists a vector in this set which
%% is maximized.
Also, if a vector $\MAT{y}$ approximates an eigenvector $\MAT{x}$ of
$\MAT{A}$, then $R(\MAT{A}, \MAT{y})$ is an approximation of
$R(\MAT{A}, \MAT{x})$. That is, eigenvalues can be approximated based
on approximations of eigenvectors by using the Rayleigh quotient.

\subsection{Spectral Extrapolation}
\label{sec:spectralextrapolation}

Recall the problem of predicting the formation in $G$ of new edges at
time $t+1$ from the adjacency matrices $\MAT{A}_1, \MAT{A}_2, \ldots,
\MAT{A}_t$ representing the evolution of $G$ over $t$ time units. The
approach is to extrapolate the trajectory of the eigenvalues of the
adjacency matrices in this sequence.

Let $(\MAT{X}_t,\MAT{\Lambda}_t)$ be the spectral decomposition of
$\MAT{A}_t$, and let $\MAT{x}_j$ and $\lambda_j$ be the $j$-th
eigenvector and eigenvalue in $\MAT{X}_t$ and $\MAT{\Lambda}_t$,
respectively (i.e., the latent dimension $j$). Under the assumption of
the spectral evolution model, the trajectory of $\lambda_j$ in
$\MAT{A}_1, \MAT{A}_2, \ldots, \MAT{A}_t$ can be approximated by
$\hat{\lambda}_{j}^{1:t}$ defined as:
\[\hat{\lambda}_{j}^{1:t} = R(\MAT{A}_1,\MAT{x}_j), R(\MAT{A}_2,\MAT{x}_j), \ldots, R(\MAT{A}_t,\MAT{x}_j).\]
Note that $\hat{\lambda}_{j}^{1:t}(t) = R(\MAT{A}_t,\MAT{x}_j) =
\lambda_j$.  The problem of computing the $j$-th eigenvalue of the
matrix $\MAT{A}_{t+1}$ (i.e., of predicting the evolution of the
spectrum for the latent dimension $j$) can then be cast as fitting the
trajectory of $\hat{\lambda}_{j}^{1:t}$. For solving this problem,
e.g., linear and quadratic regressions are available.

Computationally, the following are the calculations: obtaining the
spectral decomposition of $\MAT{A}_t$ takes $O(n^3)$ time, computing
the sequence $\hat{\lambda}_{j}^{1:t}$ takes $O(tn^2)$ (each
application of $R$ can take $O(n^2)$ time), and fitting the sequence,
e.g., $O(tn^2)$ time with least squares regression. Therefore, the
computational cost of computing $\MAT{A}_{t+1}$ can take $O(n^3 +
tn^3) = O(tn^3)$, because the spectral decomposition of $\MAT{A}_t$
can be computed just once, and there are $n$ computations taking
$O(tn^2)$ time, one per latent dimension. Note that $O(tn^3)$ is also
the asymptotic time bound for checking the spectral evolution model
hypothesis for $\MAT{A}_1, \MAT{A}_2, \ldots, \MAT{A}_t$ (namely, that
all eigenvectors of the network remain largely constant over the $t$
time units).

\begin{figure}[htbp!]
	\centering
	\begin{subfigure}[b]{0.45\textwidth}
		\centering
		\includegraphics[width=\textwidth]{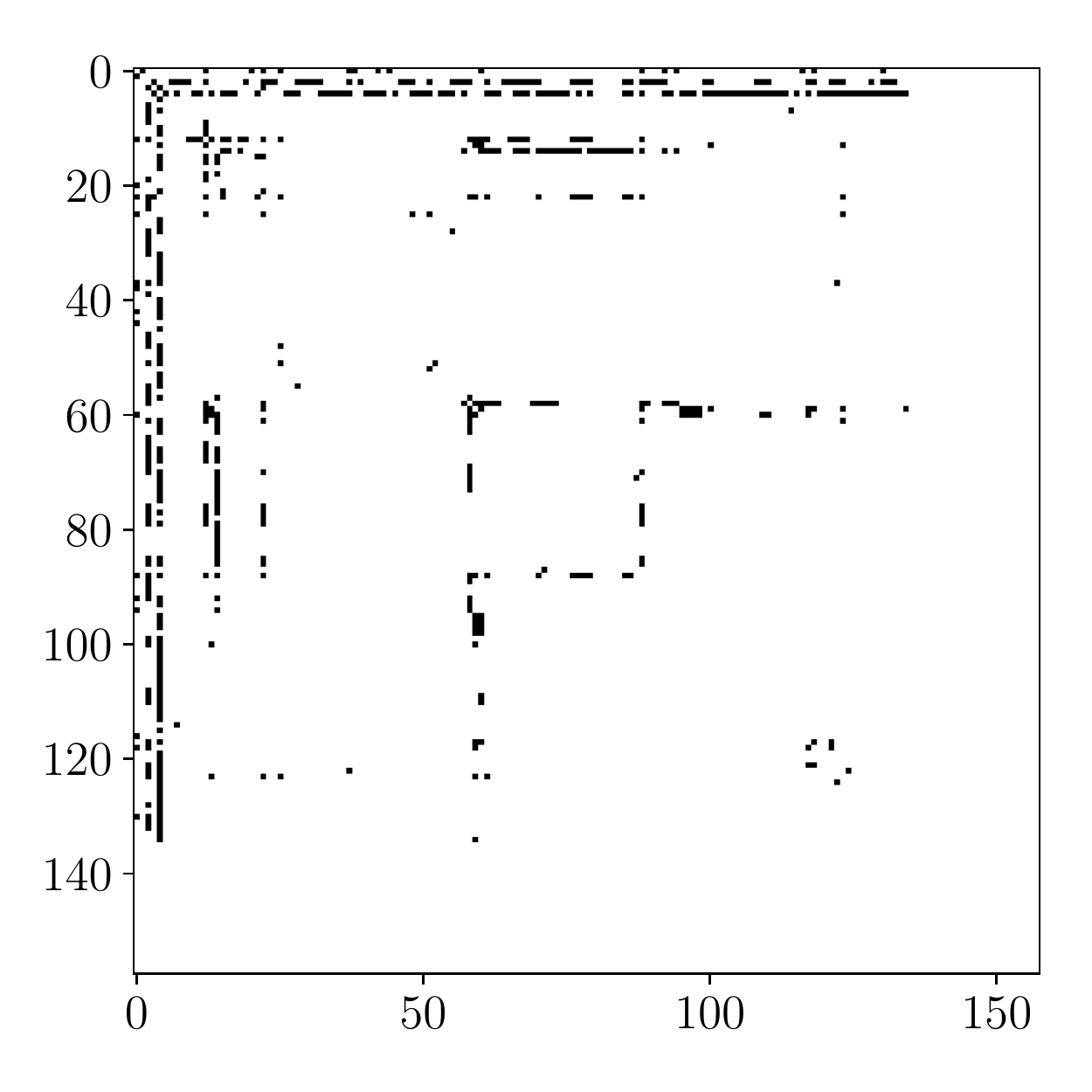}
		\caption{$\mathbf{A}_{t+1}$}
		\label{fig:adjorg}
	\end{subfigure}
	%  \qquad
	\begin{subfigure}[b]{0.45\textwidth}
		\centering
		\includegraphics[width=\textwidth]{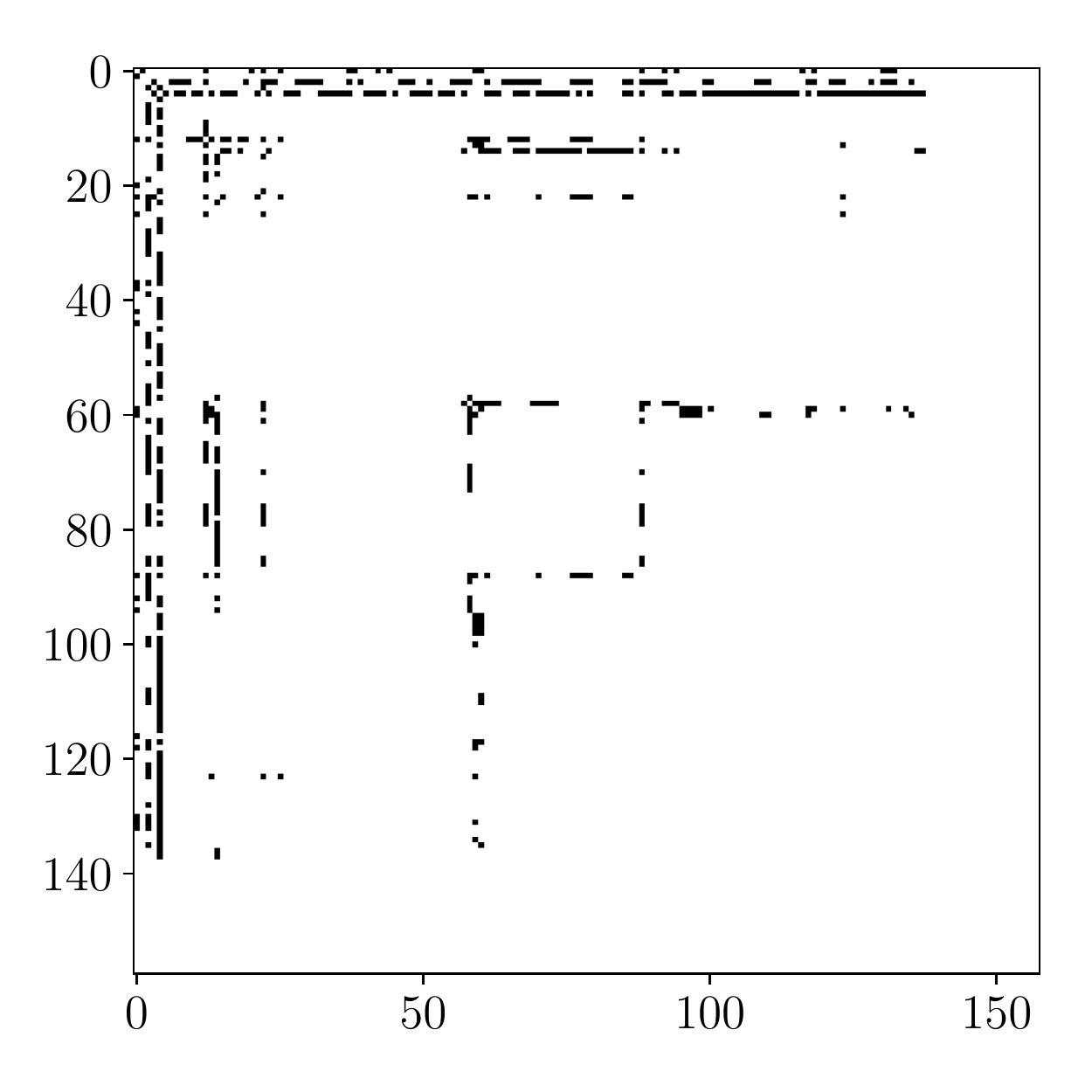}
		\caption{Reconstructed $\mathbf{A}_{t+1}$}
		\label{fig:adjrec}
	\end{subfigure}
	\caption{Adjacency matrix reconstruction of the mention network
    \#vocesdelareconciliacion. The Rayleigh quotient is used to
    approximate the eigenvalues of a matrix from a set of orthogonal
    vectors (eigenvector approximations). The accuracy of these
    approximations is showed by the comparison between the original
    matrix and its reconstruction (based on the approximations of the
    eigenvalues).}
	\label{fig:adjrq}
\end{figure}

With further experimental evaluation, the constant accompanying the
asymptotic bound $O(tn^3)$ for the construction of $\MAT{A}_{t+1}$ can
be reduced by a fraction. In particular, the approach above can be
carried out for only a selection of the eigenvectors of
$\MAT{A}_t$. For the case study presented in
Section~\ref{sec:results}, it was possible to consider about 8\% of
the eigenvectors and yet reconstruct $\MAT{A}_{t+1}$ with an accuracy
of $99.97\%$. Figure~\ref{fig:adjrq} presents the comparison between
the exact approach (i.e., extrapolating the trajectory of the $n$
eigenvalues) and the above-mentioned approach (i.e., extrapolating the
trajectory of only $8\%$ of the $n$ eigenvalues).

%% file: results.tex
\section{Case Study: Twitter Conversations}
\label{sec:results}

This section presents the results of applying the proposed growth
methods to predict evolution of the mention networks between Twitter
users. First, the networks are described and the assumptions of the
spectral evolution model are verified. Second, the performance of the
extended model is analyzed and compared to the performance of the
using common methods from the original spectral evolution model.

\subsection{Data Description}
\label{sec:data}

The dataset consists of $31$ mention networks on Twitter users with a
profile location in Colombia. These networks capture conversations
around a set of hashtags $H$ related to popular political topics
between August $2017$ and August $2018$. Users are represented by the
vertex set $V$. The set of edges is denoted by $E$. There exists an
edge $(u,v)$ between users $u$ and $v$, if $u$ identifies a message
with a political hashtag in $H$ (e.g., \#eleccionesseguras) and
mentions~$v$ (via @username). The mention network $G=(V,E)$ is a
simple graph (without self-loops), which means that there is at most
one edge between any pair of users. Moreover, every edge $(u,v)\in E$
is associated to a timestamp representing the time at which the edge
was created.

The analysis presented in this section is based on the largest
connected component of $G$, denoted by $G_c=(V_c,E_c)$. Networks $G$
and $G_c$ are built for each hashtag $h\in
H$. Table~\ref{tab:networks} presents a brief description of each
network considered in this study, including its corresponding hashtag,
and number of vertices and edges ($|V|$ and $|E|$).

\begin{table}[htbp]
	\centering
	\begin{tabularx}{\textwidth}{@{}Ll RRRR @{}}
%		\hline
		& \multirow{2}*{Set of hashtags $H$} & \multicolumn{2}{c}{\;\;\;\;\;\;\;$G$} &  \multicolumn{2}{c}{\;\;\;\;\;\;\;$G_c$}  \\
		& & $|V|$ & $|E|$ & $|V_c|$ &  $|E_c|$\\[1ex]
		\hline\hline
		1  & abortolegalya & 2235 & 2202 & 1282 & 1538\\
		2  & alianzasporlaseguridad & 176 & 1074 & 150 & 351\\
		3  & asiconstruimospaz & 2514 & 14055 & 2405 & 6950\\
		4  & colombialibredefracking & 1606 & 3483 & 1476 & 3127\\
		5  & colombialibredeminas & 707 & 2685 & 655 & 1421\\
		6  & dialogosmetropolitanos & 959 & 18340 & 932 & 4134\\
		7  & edutransforma & 166 & 1296 & 161 & 404\\
		8  & eleccionesseguras & 3035 & 17922 & 2634 & 7969\\
		9  & elquedigauribe & 2375 & 6933 & 2052 & 5272\\
		10  & frutosdelapaz & 1671 & 6960 & 1479 & 3468\\
		11 & garantiasparatodos & 388 & 814 & 340 & 563\\
		12 & generosinideologia & 639 & 914 & 615 & 805\\
		13 & hidroituangoescololombia & 1028 & 3362 & 883 & 2252\\
		14 & horajudicialur & 2250 & 23647 & 2187 & 6756\\
		15 & lafauriecontralor & 2154 & 7082 & 1999 & 5309\\
		16 & lanochesantrich & 1518 & 6946 & 1444 & 3567\\
		17 & lapazavanza & 2949 & 8288 & 2775 & 6569\\
		18 & libertadreligiosa & 1584 & 13443 & 1395 & 6856\\
		19 & manifestacionpacifica & 211 & 274 & 112 & 151\\
		10 & plandemocracia2018 & 3090 & 20955 & 2962 & 7996\\
		21 & plenariacm & 1504 & 19866 & 1460 & 4782\\
		22 & proyectoituango & 1214 & 3086 & 1186 & 1891\\
		23 & reformapolitica & 2714 & 8385 & 2608 & 5928\\
		24 & rendiciondecuentas & 5103 & 25479 & 4401 & 10308\\
		25 & rendiciondecuentas2017 & 1711 & 12441 & 998 & 2933\\
		26 & resocializaciondigna & 503 & 4054 & 496 & 1171\\
		27 & salariominimo & 2494 & 7041 & 2079 & 5016\\
		28 & semanaporlapaz & 1988 & 8103 & 1732 & 4860\\
		29 & serlidersocialnoesdelito & 530 & 861 & 439 & 697\\
		30 & vocesdelareconciliacion & 161 & 1500 & 158 & 405\\
		31 & votacionesseguras & 2748 & 13307 & 2439 & 5338\\
		\hline
	\end{tabularx}
	\caption{Description of the mention networks with political hashtags. The analysis is based on the largest connected component of each network, $G_c$.}
	\label{tab:networks}
\end{table}

\subsection{Verifying the Spectral Evolution Model Assumptions}
\label{sec:spmverif}

\begin{figure}[htbp]
  \centering
  \includegraphics[width=0.7\textwidth]{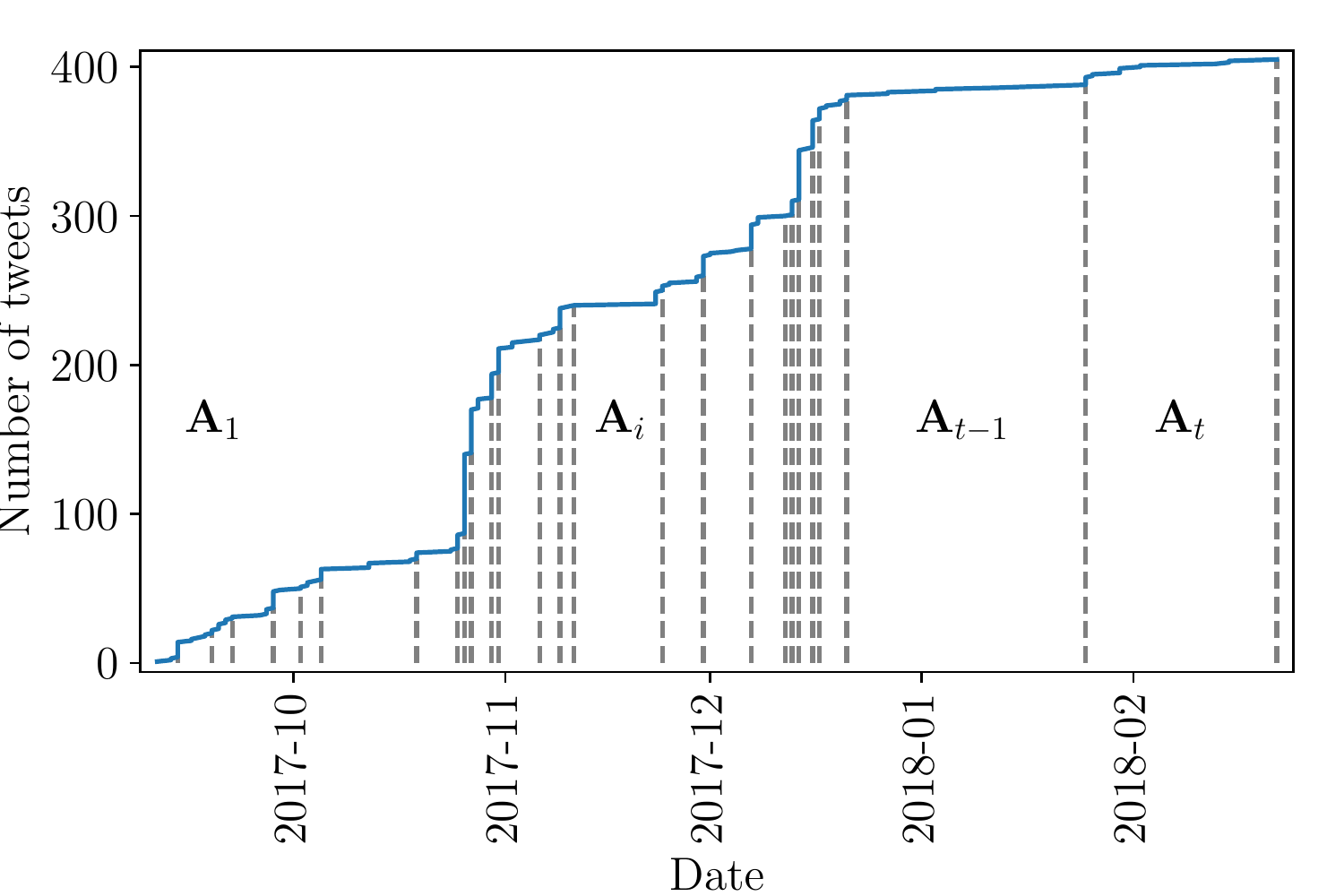}
  \caption{Process of defining a sequence of networks for the mention
    network \#vocesdelareconciliacion based on $t$ time steps. In
    general, for a network $G=(V,E)$ with adjacency matrix
    $\mathbf{A}$, the edge set $E$ is split into $t$ disjoint subsets
    according to their associated time stamps. Time step $i$ contains
    all the edges of the first $i$ subsets of edges. Note that each
    time $i$ is associated to an adjacency matrix $\mathbf{A}_i$ and
    $\mathbf{A}_t=\mathbf{A}$.}
  \label{fig:steps}
\end{figure}

To apply the original and extended versions of the spectral evolution
model, the assumptions on the spectrum and eigenvectors need to be
verified.  Let $t \in \nats$ be a positive number. The set of edges
$E$ is sorted by time stamps and then split into $t$ disjoint subsets
of equal size ($|E|/t$). Next, consider a total of $t$ time steps,
created in a way that each time $i$, $1\leq i\leq t$, contains the
edges of the first $i$ subsets of edges. Note that each time step $i$
represents a cumulative graph that is associated to an adjacency
matrix $\mathbf{A}_i$ and its spectral decomposition
$\mathbf{X}_i\mathbf{\Lambda}_i\mathbf{X}^T_i$.  Note also that
$\mathbf{A}_1,\mathbf{A}_2,\dots,\mathbf{A}_t$ represents the
evolution of the network over time and $\mathbf{A}_t=\mathbf{A}$
represents the adjacency matrix of the complete network.  Overall,
this process takes $O(tn^3)$ time with $n = |V|$:
each spectral decomposition takes $O(n^3)$
and there are $t$ of such decompositions to compute.
Figure~\ref{fig:steps} summarizes the process of defining a sequence
of networks for a particular mention network.

\subsubsection{Spectral and Eigenvector Evolution}

\begin{figure}[htbp]
  \centering
  \begin{subfigure}[b]{0.5\textwidth}
    \centering
    \includegraphics[width=\textwidth]{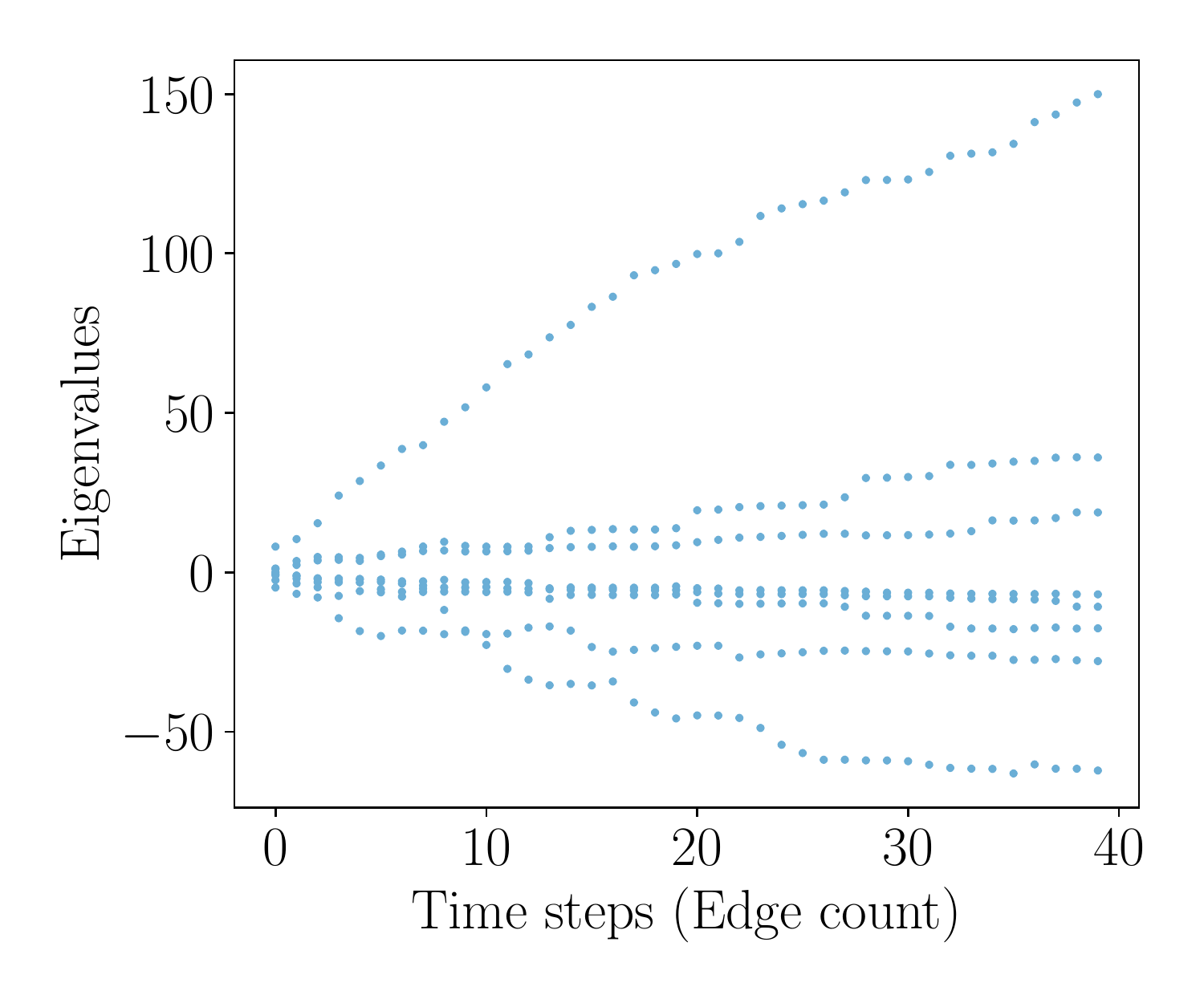}
    %    \caption{\#EduTransforma}
    \label{fig:specevol1}
  \end{subfigure}
  \hspace{-0.4cm}
  \begin{subfigure}[b]{0.5\textwidth}
    \centering
    \includegraphics[width=\textwidth]{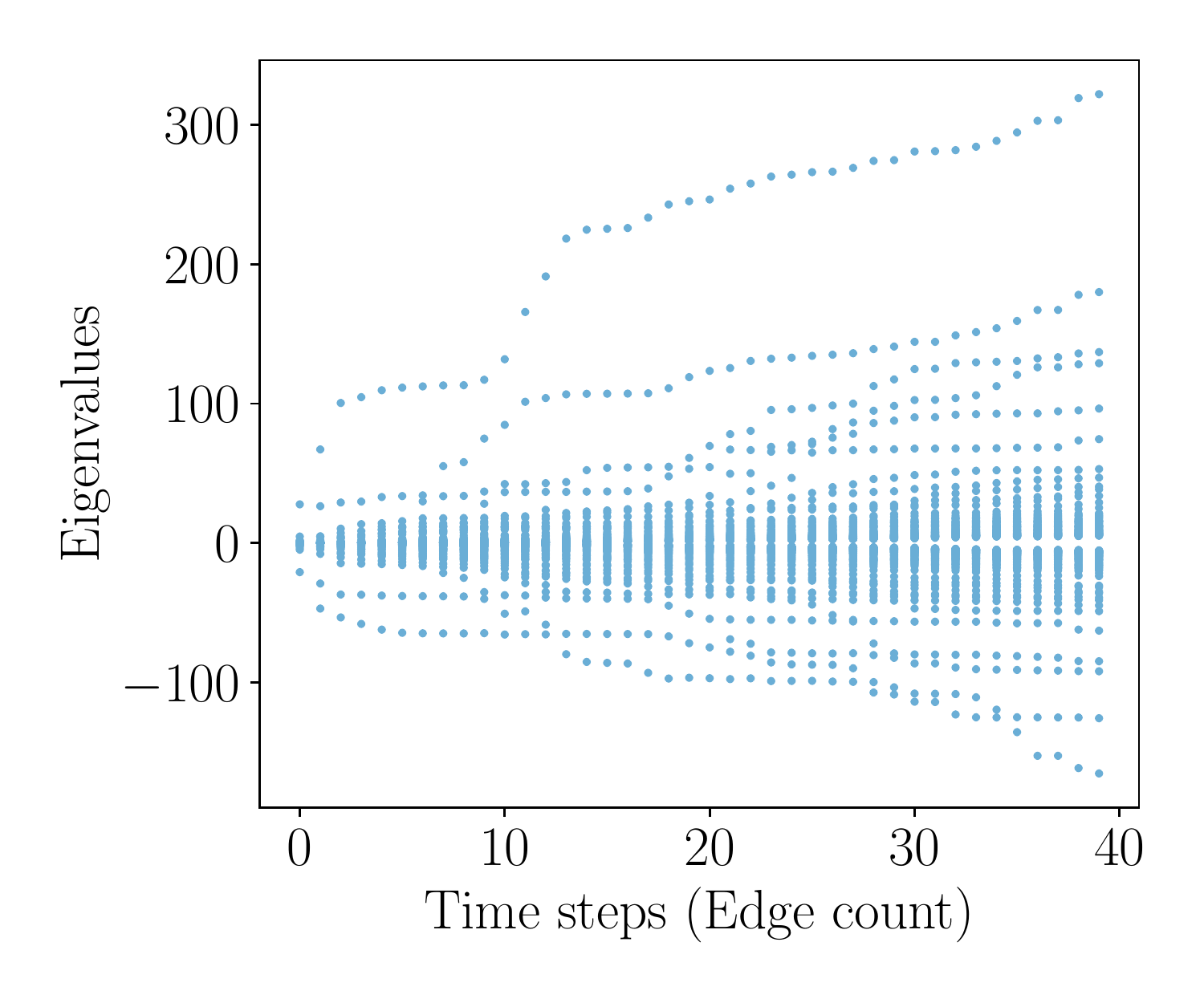}
    %    \caption{\#AsiConstruimosPaz}
    \label{fig:specevol2}
  \end{subfigure}
  \vspace{-0.6cm}
  \caption{Spectral evolution for two mention networks,
    \#edutransforma (left) and \#asiconstruimospaz (right). Plots
    illustrate the top $8\%$ of the largest eigenvalues by absolute
    value at each time step.}
  \label{fig:specevol}
\end{figure}

Figure~\ref{fig:specevol} identifies the largest eigenvalues (by
absolute value) for two mention networks: namely,
\#educationtransforms and \#howwebuildpeace.  Note that for each of
these two networks, the spectra grows irregularly.  Consider the
adjacency matrix $\mathbf{A}_1$ for time $t_1$ satisfying $1\leq
t_1<t$. The eigenvectors ${(\mathbf{X}_t)}_j$ corresponding to the
largest eigenvalues are compared to the eigenvectors
${(\mathbf{X}_1)}_j$at time $t$ using the cosine distance as a
similarity measure, for each latent dimension $j$.

%\vspace{-1cm}

\begin{figure}[htbp]
  \centering
  \begin{subfigure}[b]{0.5\textwidth}
    \centering
    \includegraphics[width=\textwidth]{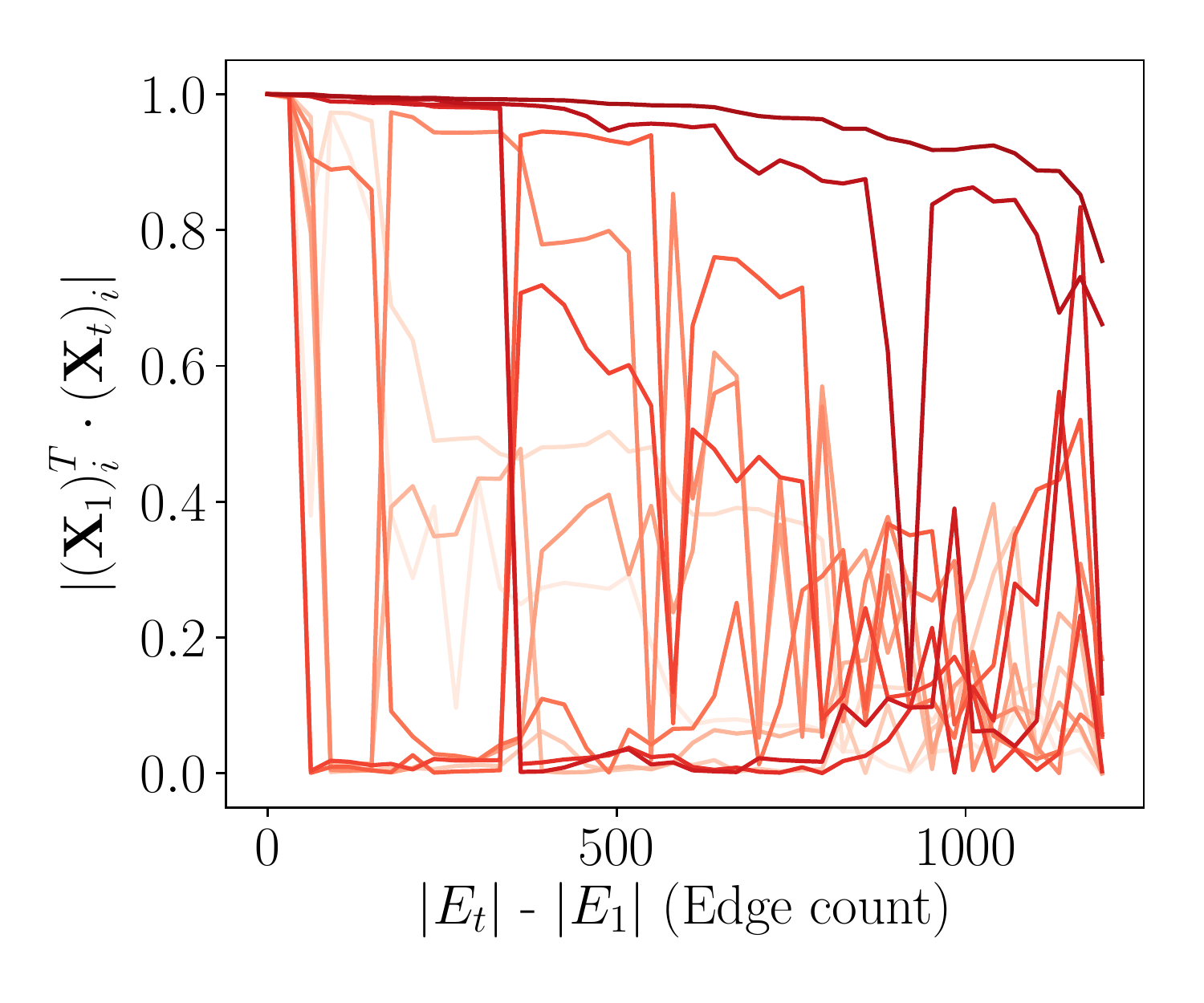}
    %\caption{\#VocesDeLaReconciliacion}
    \label{fig:vecevol1}
  \end{subfigure}
  \hspace{-0.45cm}
  %  \qquad
  \begin{subfigure}[b]{0.5\textwidth}
    \centering
    \includegraphics[width=\textwidth]{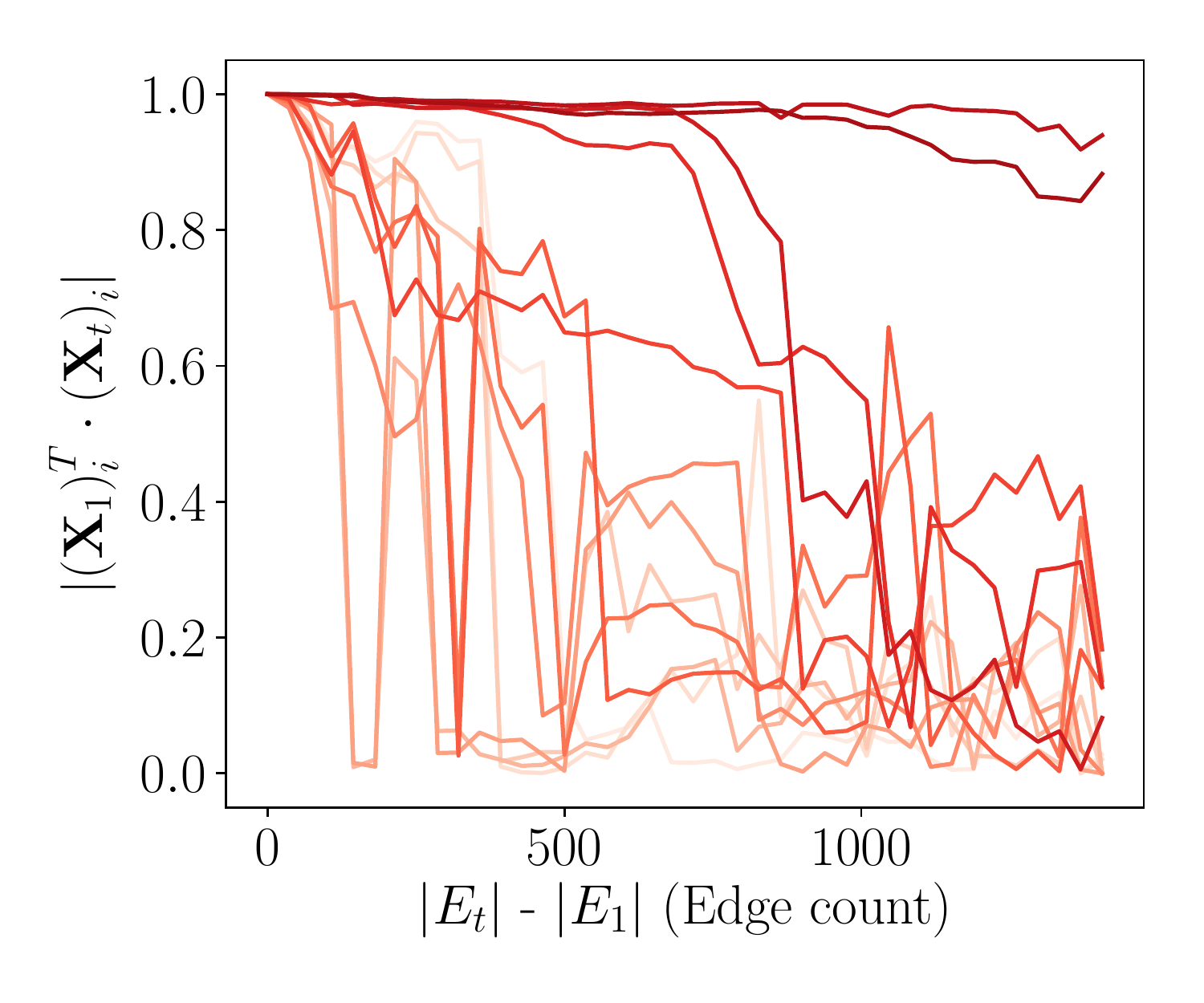}
    %\caption{\#VozPopuliTeVe}
    \label{fig:vecevol2}
  \end{subfigure}
  \vspace{-0.6cm}
  \caption{Eigenvector evolution shows a one by one eigenvector
    comparison between $t_1$ and $t$ using the cosine similarity. Each
    line represents a latent dimension, i.e., a pair of eigenvectors
    being compared, with brighter colors for eigenvectors
    corresponding to the largest eigenvalues. Eigenvector evolution for
    two mention networks, \#edutransforma (left) and
    \#vocesdelareconciliacion (right).}
  \label{fig:vecevol}
\end{figure}

Figure~\ref{fig:vecevol} shows that, during the evolution of the
network, certain eigenvectors have a similarity value close to
one. They correspond to the eigenvectors associated to the largest
eigenvalues. Note also that at points the similarity between some
eigenvectors drops to zero, which can be explained by eigenvectors
changing their order during the spectral decomposition.

%\vspace{-1cm}

\subsubsection{Eigenvector Stability}
\label{sec:eigenstab}

\begin{figure}[htbp!]
  \centering
  \includegraphics[width=0.6\textwidth]{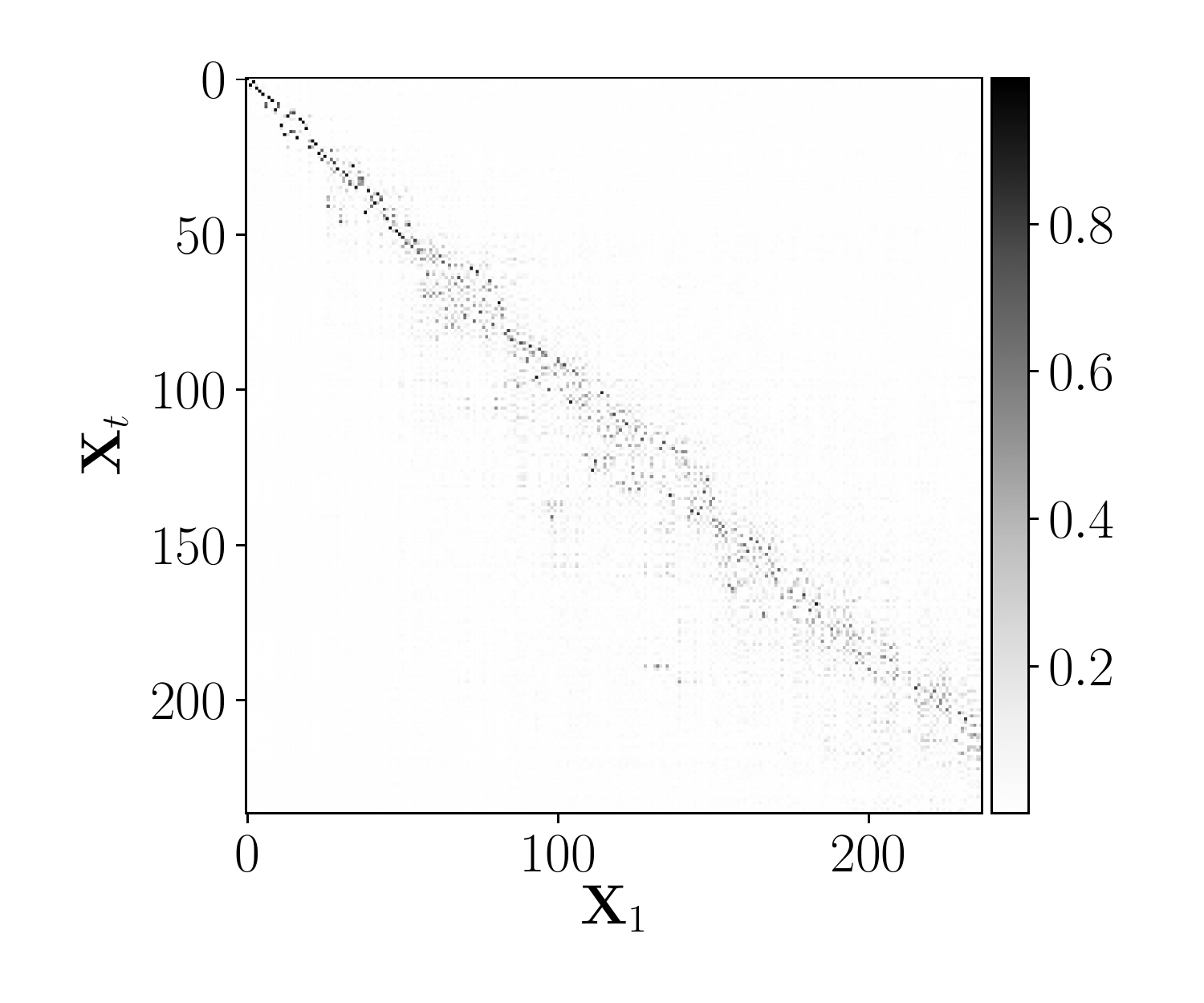}
  \vspace{-0.5cm}
  \caption{Eigenvector stability for the mention network
    \#plandemocracia2018. Similarity measure for every pair of
    eigenvectors between $t_1$ and $t$. A similarity value of zero is
    represented by a white square and a similarity of one by a black
    square. The matrix illustrates the permutations of the
    eigenvectors. A diagonal matrix represents the case where
    eigenvectors are preserved over time.}
  \label{fig:stab}
\end{figure}

Let $t_1$ and $t$ denote the times at which $75\%$ and $100\%$ of all
edges in a given network $G$ are present in its adjacency matrices
$\MAT{A}_1$ and $\MAT{A}_t$ at times $t_1$ and $t$,
respectively. Their spectral decomposition is
$\mathbf{A}_1=\mathbf{X}_1\mathbf{\Lambda}_1\mathbf{X}^T_1$ and
$\mathbf{A}_t=\mathbf{X}_t\mathbf{\Lambda}_t\mathbf{X}^T_t$. Similarity
values are computed for every pair of eigenvectors $\MAT{x}_i$ and
$\MAT{x}_j$ by:
\[\operatorname{sim}_{ij}(t_1,t)=\left|{(\mathbf{X}_1)}^T_i\cdot{(\mathbf{X}_t)}_j\right|.\]
Figure~\ref{fig:stab} summarizes the resulting values, plotted as a
heatmap. White cells represent a value of zero and black cells a value
of one. The more the heatmap approximates a diagonal matrix, the fewer
eigenvector permutations there are (i.e., the eigenvectors are
preserved over time). Note that the value of certain cells in
Figure~\ref{fig:stab} is between zero and one. These cells result
either from an exchange in the location of eigenvectors that have
eigenvalues very close in magnitude or from comparing eigenvalues
${({\lambda}_1)}_i$ and ${({\lambda}_t)}_i$ that do not correspond to
each other.

\subsubsection{Spectral Diagonality Test}

\begin{figure}[htbp]
  \centering
  \includegraphics[width=0.55\textwidth,center]{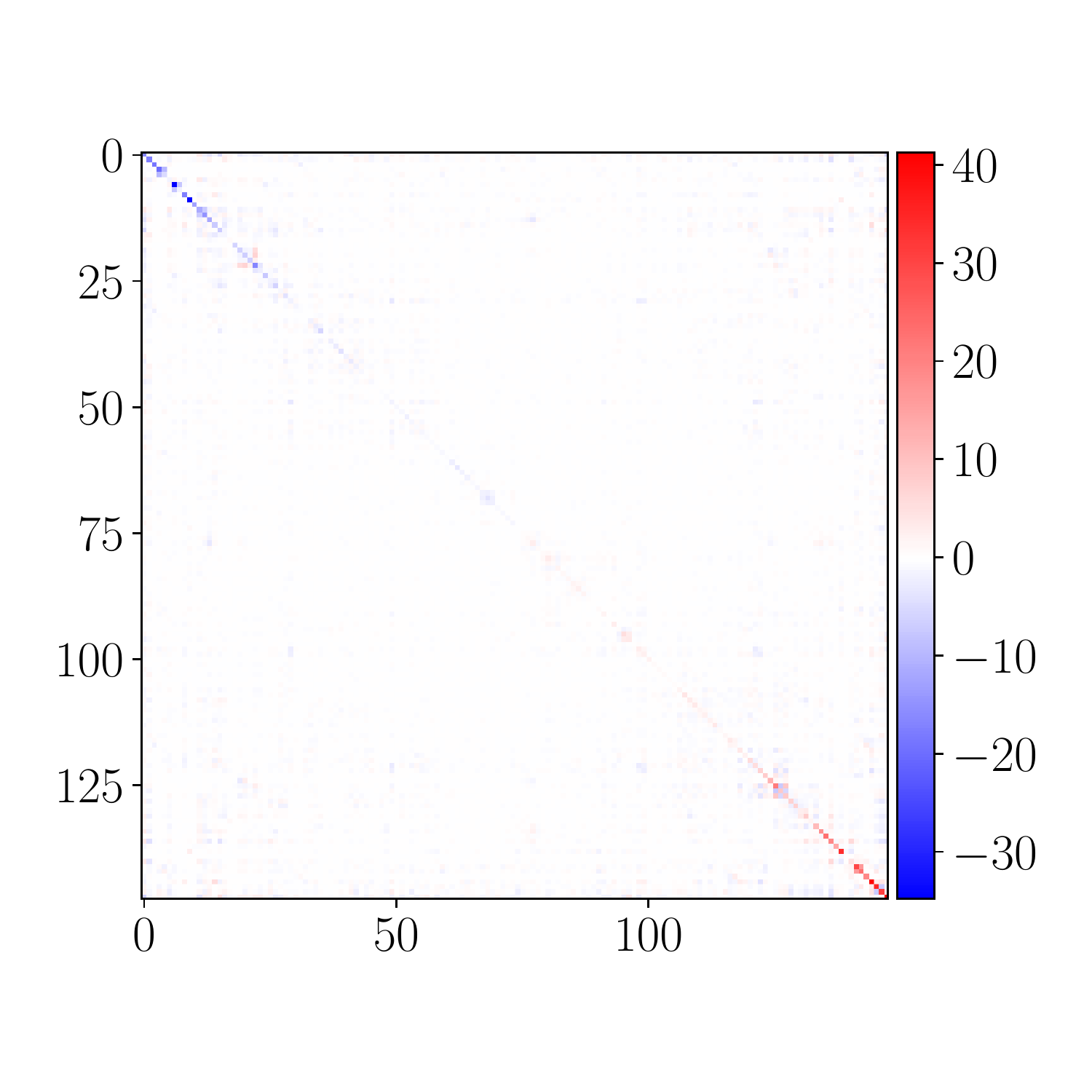}
  \vspace{-0.5cm}
  \caption{The spectral diagonality test shows to what extent the
    network evolution is explained by the eigenvalues. If network
    growth is related to spectrum evolution, then the matrix should be
    approximately diagonal.  Spectral diagonality test for the mention
    network \#plandemocracia2018.}
  \label{fig:diag}
\end{figure}

Recall the spectral decomposition $(\MAT{X}_1, \MAT{\Lambda}_1)$ of
$\MAT{A}_1$ from Section~\ref{sec:eigenstab}, representing the network
$G$ at time $t_1$. At any time $t_2>t_1$, the adjacency matrix
$\MAT{A}_2$ is expected to have the form
$\mathbf{A}_2=\mathbf{X}_1(\mathbf{\Lambda}_1 +
\mathbf{\Delta})\mathbf{X}^T_1$, where $\mathbf{\Delta}$ is a diagonal
matrix. The diagonal matrix $\mathbf{\Delta}$ can be used to infer if
the network's growth is related to the eigenvalues. Technically, by
using least-squares, the matrix $\mathbf{\Delta}$ can be derived as
$\mathbf{\Delta}=\mathbf{X}_1(\mathbf{A}_2-\mathbf{A}_1)\mathbf{X}^T_1$. If
$\mathbf{\Delta}$ is diagonal, then the growth between $t_1$ and $t_2$
is spectral.  Figure~\ref{fig:diag} shows the diagonality test for the
mention network \#plandemocracia2018. It has been found that most of
the networks in the dataset show irregular behavior in the spectrum
evolution. Note that the matrix $\mathbf{\Delta}$ (spectral
diagonality) is almost diagonal for all mention networks.

\subsection{Prediction Performance}
\label{sec:perform}

\begin{figure}[htbp]
  \centering
  \includegraphics[width=.75\textwidth]{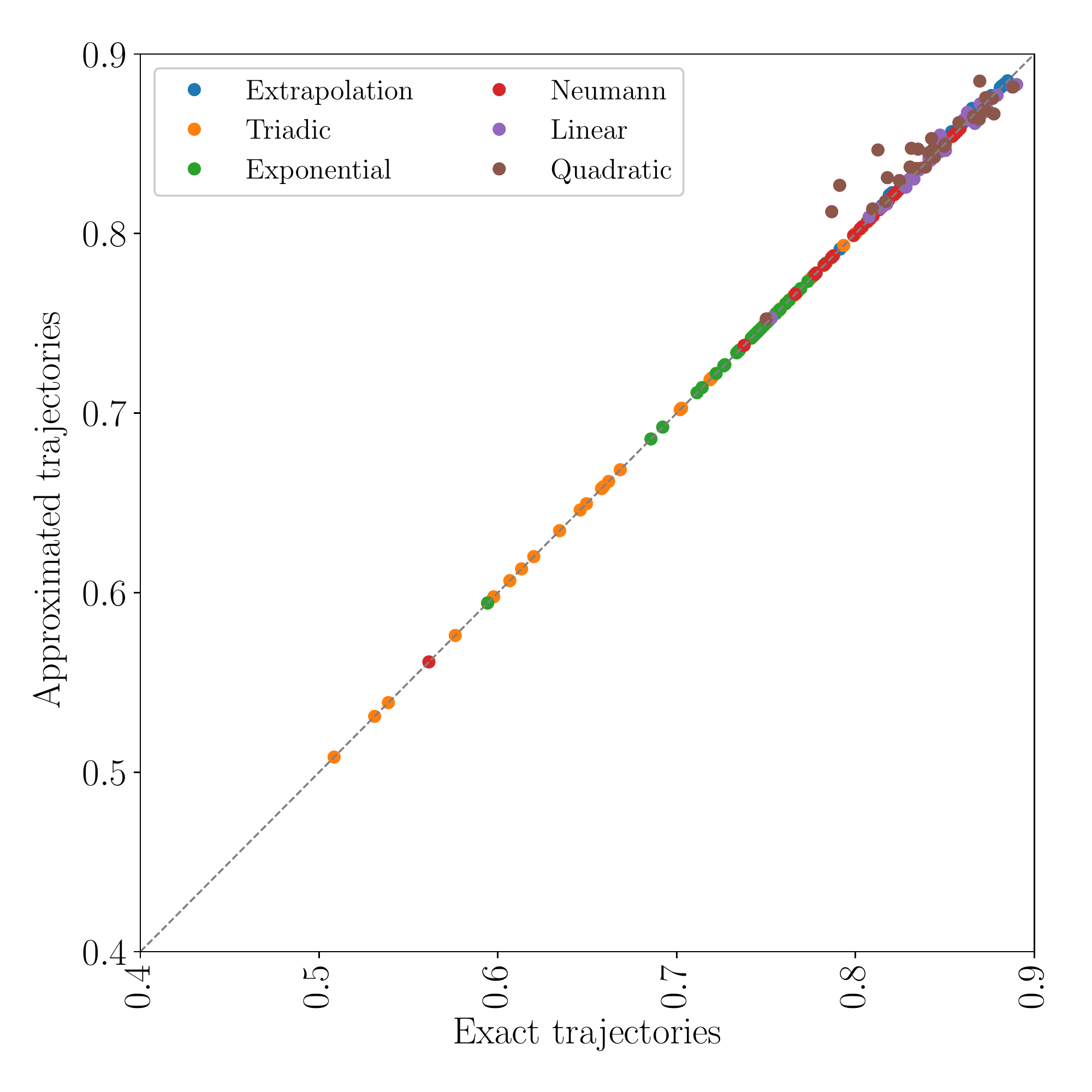}
  \caption[Performance evaluation]{Performance comparison of using
    exact spectral decomposition and the proposed approach where
    spectrum trajectories are approximated using the Rayleigh
    quotient. The comparison is measured by the AUC ROC score. Both
    methods seem to be consistent, i.e., the performance of the
    prediction is replicated using the Rayleigh quotient method.}
  \label{fig:aucroc75}
\end{figure}

First, the prediction performance of the proposed approach
(Section~\ref{sec:extmodel}) is compared to that of the spectral
evolution model (Section~\ref{sec:model}). In each case, the
performance of the prediction is measured by the AUC ROC score.
Figure~\ref{fig:aucroc75} summarizes the performance comparison of the
growth methods when exact spectral decomposition and trajectories
approximations with the Rayleigh quotient. Note that for most of the
networks and growth methods, the performance of the two approaches is
highly similar. One conclusion to draw is that it is possible to
replicate the prediction of the exact method with the more efficient
method based on approximations. A $75\%$-$25\%$ train-test split was
used for the analysis depicted in Figure~\ref{fig:aucroc75}. This
ratio can be interpreted as $75\%$ of edges of the network (by
creation time) are used to predict the remaining $25\%$. The proposed
approach is used for the remainder of the analysis in this section.

\begin{figure}[htbp]
  \centering
  \includegraphics[width=.75\textwidth]{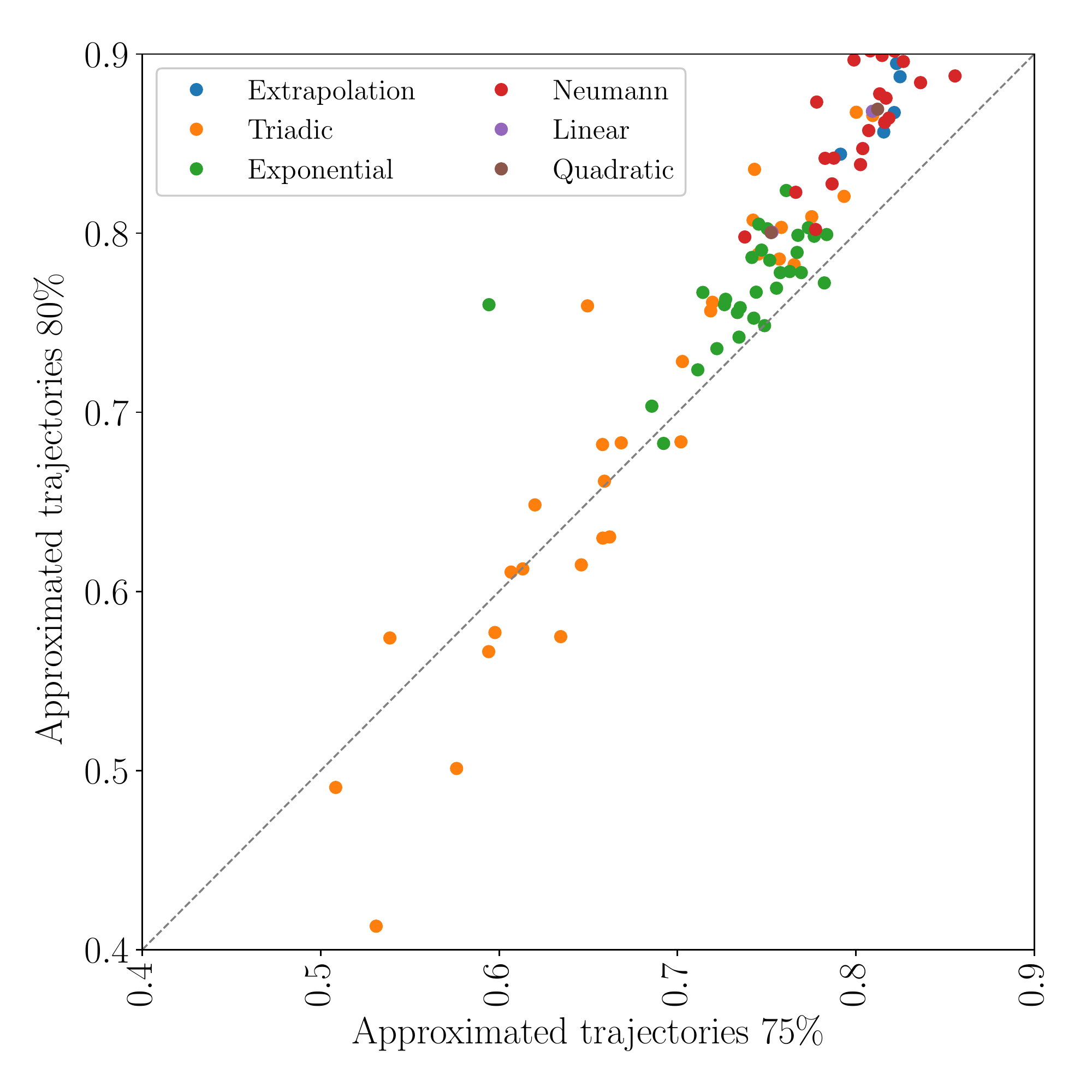}
  \caption[Performance evaluation]{Comparison of the prediction
    performance based on AUC ROC score for $75\%$ and $80\%$
    train-test ratios. Increasing the train set size improves the
    overall performance of the growth methods.}
  \label{fig:aucall}
\end{figure}

%% Given that $t=40$, $25\%$ of edges are equivalent to several time
%% steps (more than one time step is being predicted at once).
Reducing the number of steps to be predicted while avoiding
overfitting, can bring performance improvement of the
prediction. Figure~\ref{fig:aucall} shows a comparison of the
performance for the $75\%$ and $80\%$ split ratios.  Note that
reducing the size of test set (and increasing train set) comes with
noticeable improvement in the performance of the methods, in
general. It has been verified that the growth of the spectrum is
irregular for most of the networks. Therefore, it is expected that
methods based on extrapolation, and linear and quadratic regression,
outperform the ones based on graph kernels.

\begin{figure}[htbp]
  \centering
  \includegraphics[width=.6\textwidth]{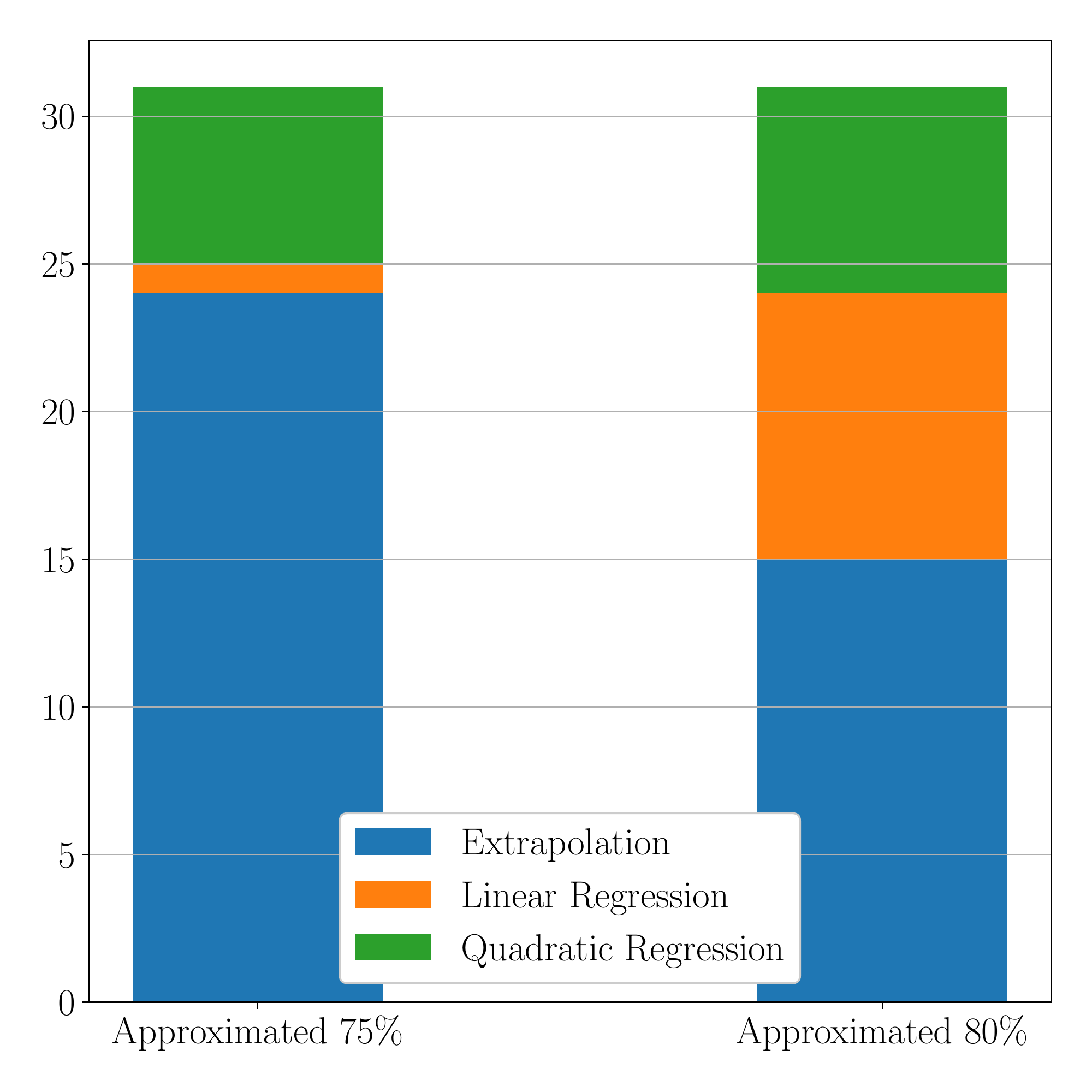}
  \caption[Performance evaluation]{With a $75\%$-$25\%$ train-test
    ratio for the prediction, the extrapolation method outperforms the
    regression methods for 24 out of the 31 analyzed networks. When
    the train set size increases to $80\%$, the regression methods
    outperform the extrapolation methods for 16 networks.}
  \label{fig:aucbest}
\end{figure}

\begin{figure}[htb!]
  \centering
  \includegraphics[width=.85\textwidth]{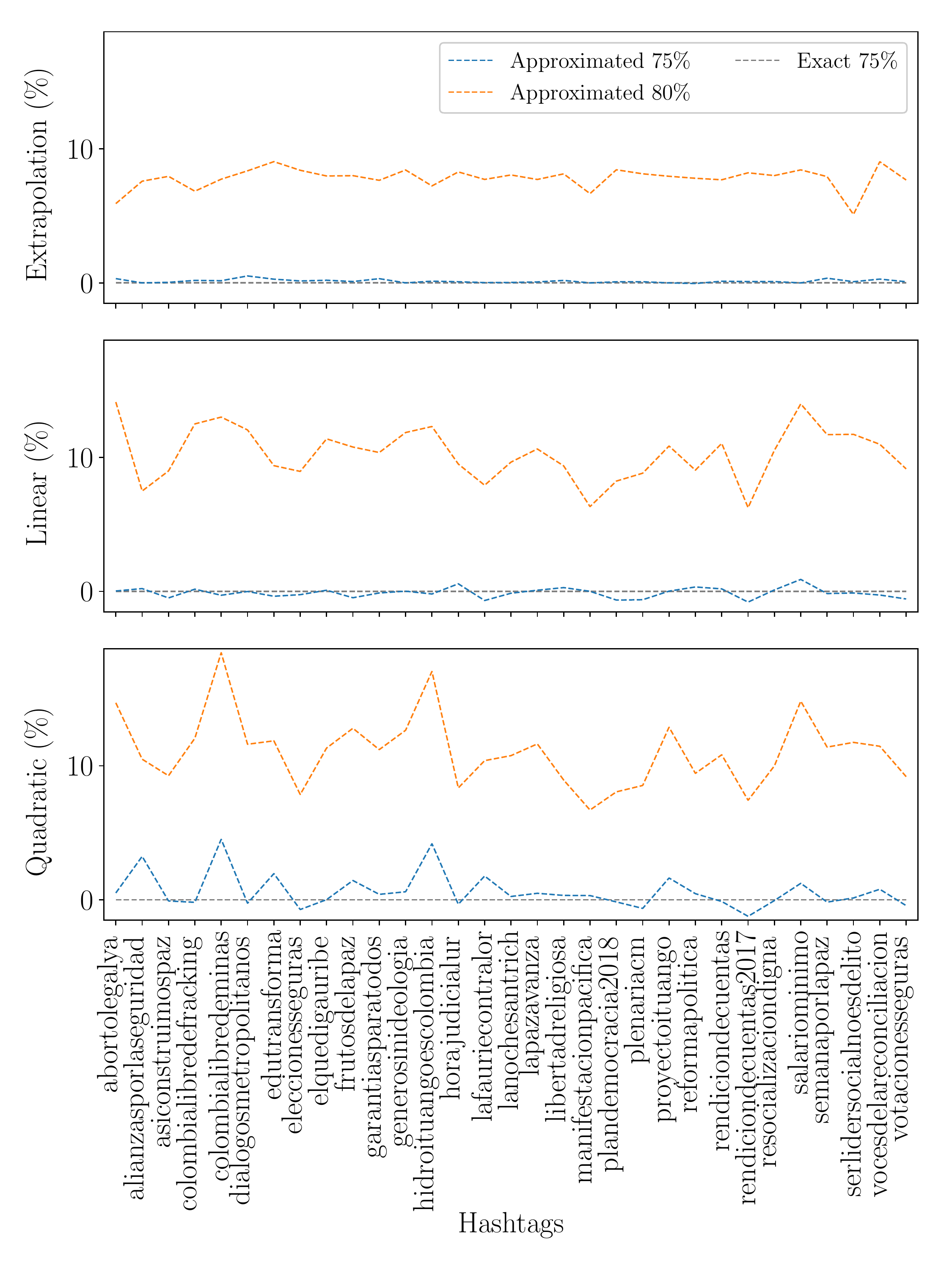}
  \caption[Performance evaluation]{Improvement of the AUC ROC score in
    percentage computed with the Rayleigh quotient considering
    different train-test ratios for extrapolation (top), linear
    (middle) and quadratic (bottom) growth methods w.r.t. the spectral
    decomposition method. Performance of the regression methods
    improve more than $10\%$ on average.}
  \label{fig:aucper}
\end{figure}

It is important to include the AUC ROC score to evaluate the
methods. For instance, if the initial train-test ratio for both
methods is considered, the extrapolation method seems to outperform
the other methods in many cases. Its score is the best 24 times in
total. But the situation changes when the size of the training set
size increases: if the ratio is $80\%$, the regression methods get
better score for 16 out of 31 networks (9 and 7 for the linear and
quadratic regressions, respectively). As depicted in
Figure~\ref{fig:aucbest}, the distribution of the best method changes
alongside the modification of the train-test ratio. Improvement in the
regression methods is explained by the addition of new data available
for the prediction.  Figure~\ref{fig:aucper} shows the improvement of
the AUC ROC score in percentage with respect to the $75\%$-$25\%$
ratio.

\begin{figure}[htbp]
  \centering
  \includegraphics[width=.85\textwidth]{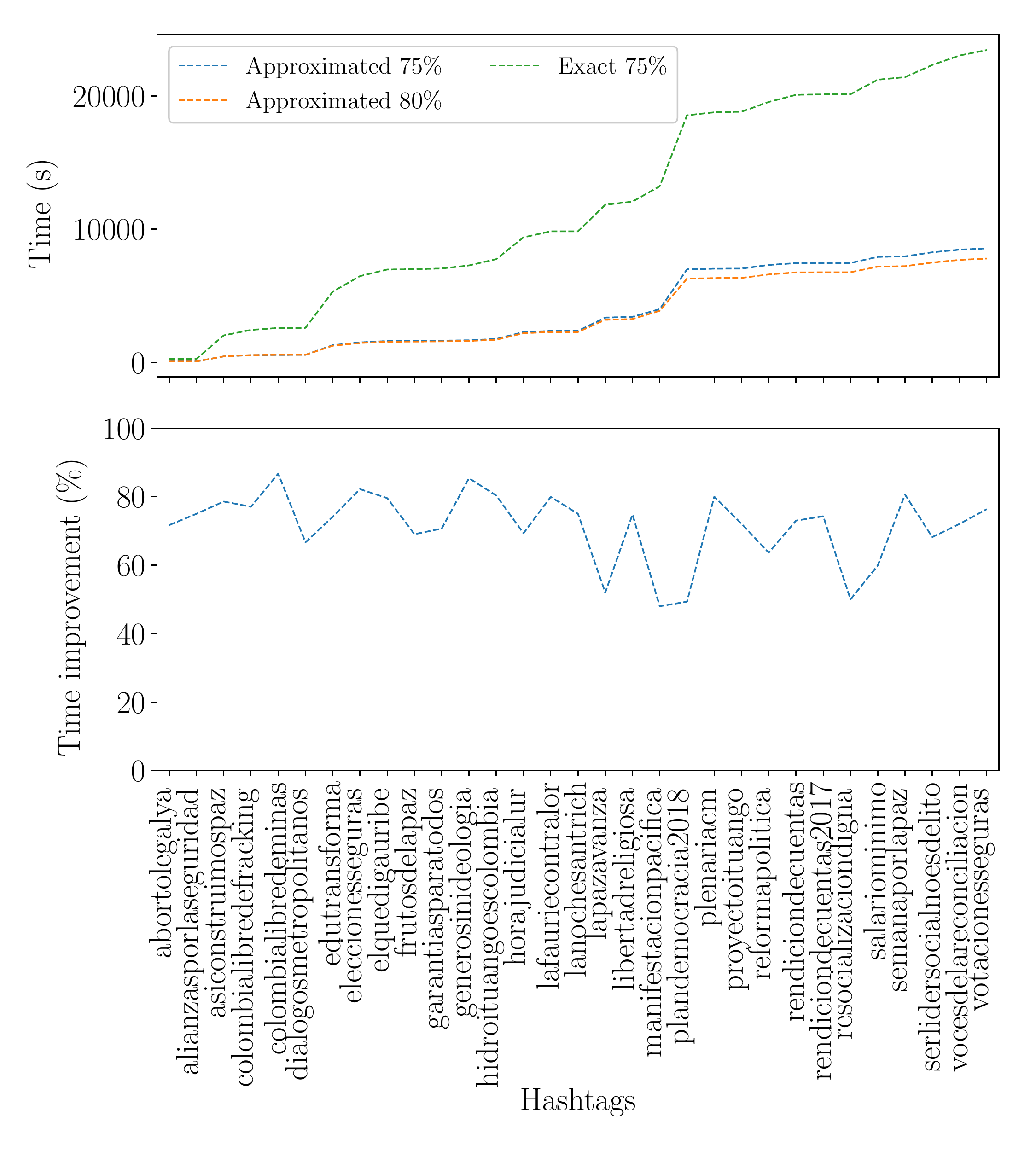}
  \caption[Performance evaluation]{Cumulative execution time of using
    the spectral evolution model with exact spectral decomposition
    ($75\%$) and the extended spectral evolution model with
    approximated spectra ($75\%$, $80\%$) when used on all 31 mention
    networks (top) and average execution time improvement (bottom).}
  \label{fig:auctime}
\end{figure}

Finally, execution times of both approaches are analyzed.
Figure~\ref{fig:auctime} (top) shows the cumulative execution time of
using the spectral evolution model with exact spectral decomposition
($75\%$) and the extended spectral evolution model with approximated
spectra ($75\%$, $80\%$) when used on all 31 mention networks. Note
that there is noticeable difference (more than $\numprint{10000}$
seconds) between the execution time of the two approaches. Considering
the difference between the methods for each network, the average
improvement in favor of the extended evolution model is $71.5\%$.  The
bottom plot in Figure~\ref{fig:auctime} summarizes the execution time
improvement in percentage for each network, ranging from $45\%$ to
$80\%$. Note that the spectral extrapolation, linear and quadratic
regression outperform the graph kernels. Better prediction in the
extrapolation-based approaches seems to be consequence of their
ability to capture the irregular evolution of the eigenvalues. Also,
linear and quadratic regressions improve their performance when the
size of the train test increases.

%% file: concl.tex
\section{Conclusion and Future Work}
\label{sec:concl}

This paper extended the spectral evolution
model~\citep{kunegis-semodel-2013} to estimate link prediction in
complex networks. It characterizes the evolution of a network in terms
of the evolution of its spectrum, i.e., the eigenvalues of its
adjacency matrix. On the one hand, the proposed approach uses the
Rayleigh quotient for fast computation of eigenvalue
approximations. On the other hand, the estimation of new edges
formation is cast as the problem of predicting eigenvalues'
trajectories. In particular, linear and quadratic regression methods
showcase the proposed approach.  It has been validated that in the
growth of some Twitter mention networks, the eigenvectors remain
largely constant, while their spectrum grows irregularly. Since
computing the spectral decomposition at every time step for each
network is a computationally expensive task, the Rayleigh quotient
proved to be a feasible tool to efficiently calculating the
approximated eigenvalues trajectories. It also avoids the exchange in
the location of eigenvectors seen in stability tests.
It was further observed that the prediction performance of some
methods improve by modifying the train-test ratio in such a way that
the size of the train set increases. Especially, the performance of
the linear and quadratic regression methods are enhanced because the
ratio modifications include more available information for the
prediction. Also, the execution time reduction introduced by the
Rayleigh quotient is noteworthy. In fact, the analysis using this
technique is twice as fast, in comparison to other common techniques,
for most of the Twitter mention networks studied here. For
experimentation, this was crucial because accessing large networks
became an option. Based on the extensive experimental exploration
summarized in this manuscript, the authors strongly believe that
proposed approach has the potential to become a significant addition
to the spectral evolution model approach.

For future research, the use of matrix representations other than the
adjacency matrix need to be researched. In general, it is known that
eigenvalues and eigenvectors are most meaningful when used to
understand a natural operator or a natural quadratic form, such as the
Laplacian matrix. In applications, it would be interesting to apply
the approach to other networks such as gene co-expression networks,
where genes represent vertices and there is a link between two genes
$g_1$ and $g_2$ if there is a positive relation in their expression,
i.e., $g_1$ and $g_2$ use to express at the same
time~\citep{ruan-coexp-2010,stuart-coexp-2003a}. In this case, new
relations in the gene co-expression can be predicted through the
extended spectral evolution model. Finally, considering different
growth methods in the analysis in order to capture more information
from the spectrum of the networks remains an important research
direction.